\documentclass[10pt,twocolumn,letterpaper]{article}

\usepackage{cvpr}
\usepackage{times}
\usepackage{epsfig}
\usepackage{graphicx}
\usepackage{amsmath}
\usepackage{amssymb}

\usepackage{color}
\usepackage{floatrow}
\usepackage{graphicx,subfigure}
\usepackage{xcolor,colortbl}

\usepackage[pagebackref=true,breaklinks=true,letterpaper=true,colorlinks,bookmarks=false]{hyperref}

\cvprfinalcopy 

\def\cvprPaperID{4153} 

\ifcvprfinal\fi
\begin{document}
	
	\title{What Object Should I Use? - Task Driven Object Detection}
	
	\author{Johann Sawatzky\thanks{contributed equally, alphabetically ordered}
		\qquad
		Yaser Souri\footnote[1]
		{\tt\small souri@iai.uni-bonn.de}
		\qquad
		Christian Grund
		\qquad
		Juergen Gall\\
		University of Bonn \\ {\tt\small \{jsawatzk, ysouri, grund, jgall\} @ uni-bonn.de}
	}
	
	\maketitle
	
	\begin{abstract}
		When humans have to solve everyday tasks, they simply pick the objects that are most suitable.
		While the question which object should one use for a specific task sounds trivial for humans, it is very difficult to answer for robots or other autonomous systems.
		This issue, however, is not addressed by current benchmarks for object detection that focus on detecting object categories.
		We therefore introduce the COCO-Tasks dataset which comprises about 40,000 images where the most suitable objects for 14 tasks have been annotated. We furthermore propose an approach that detects the most suitable objects for a given task. The approach builds on a Gated Graph Neural Network to exploit the appearance of each object as well as the global context of all present objects in the scene. In our experiments, we show that the proposed approach outperforms other approaches that are evaluated on the dataset like classification or ranking approaches.
		
	\end{abstract}
	
	\section{Introduction}

	The task of object detection in images has been widely studied and the community achieved impressive progress
	on datasets like COCO \cite{lin2014microsoft} or Pascal VOC \cite{Pascal}. For many applications like assistive or autonomous systems, however, it is insufficient to detect all instances of a set of object categories. Similar to humans, the systems interact with the environment to solve certain tasks. For instance, if a service robot is asked to serve a glass of wine, detecting all glasses in an image does not answer the question which of them it should use. Taking a beer glass is definitely the wrong choice if a wine glass is available, but if no other glasses are available it might be the best option for the task. Even if there are several wine glasses, not all of them are necessary suitable since some of the glasses might be already used by someone else or need to be cleaned. If no glasses are available, some alternatives have to be considered. For instance, wine can be drunk from a cup or jug as well. This shows that answering the question, which object should be used for a task is very difficult since it depends on the present object categories in an image and the properties of the objects.
	
	\begin{figure}[t]\label{fig:teaser}
		\centering
		\subfigure{\includegraphics[height=30mm]{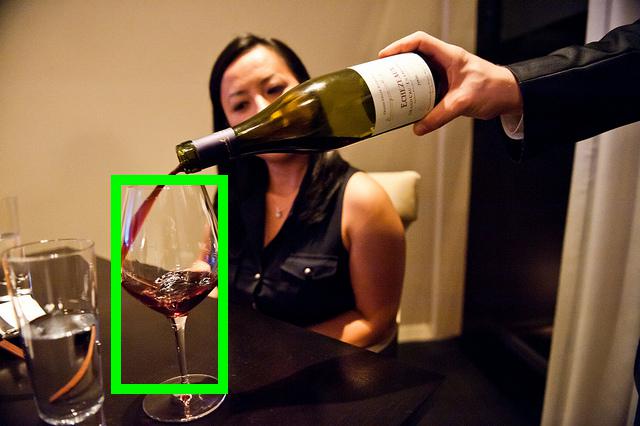}}
		\subfigure{\includegraphics[height=30mm]{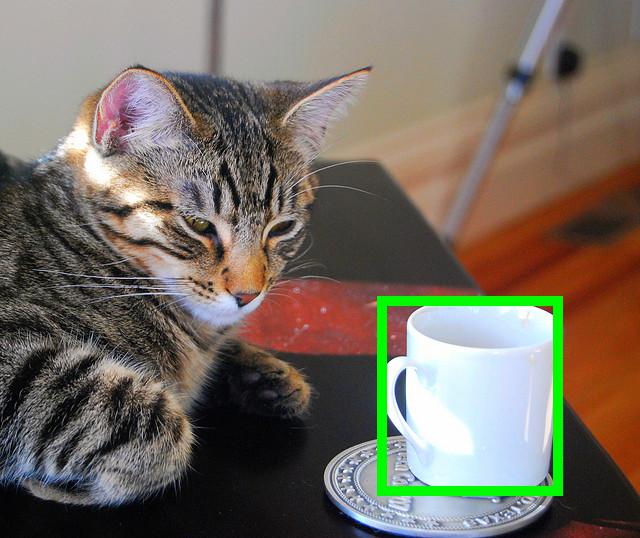}}
		\caption{What object in the scene would a human choose to serve wine? In the left image, the wine glass is preferred to other drinking glasses. In the right image, neither a wine glass nor other drinking glasses are present. The cup is therefore chosen by the human.
		}
	\end{figure}

\begin{figure*}
		\subfigure{\includegraphics[height=38mm]{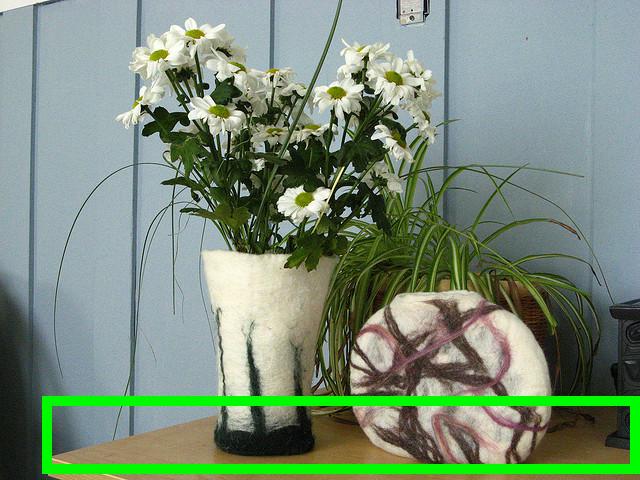}}~~
		\subfigure{\includegraphics[height=38mm]{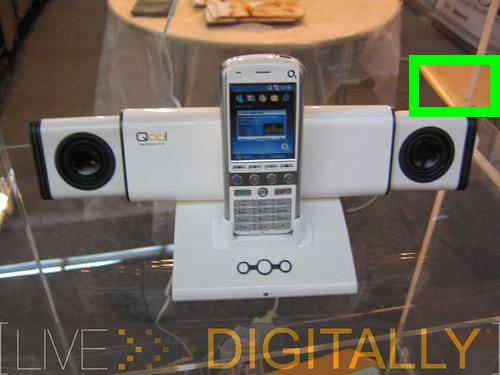}}~~
		\subfigure{\includegraphics[height=38mm]{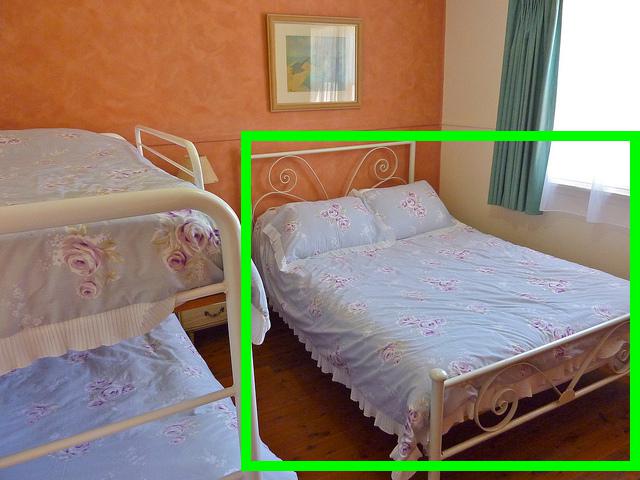}}~~\\
		\vspace{-0.2cm}
		\subfigure{\includegraphics[height=40.5mm]{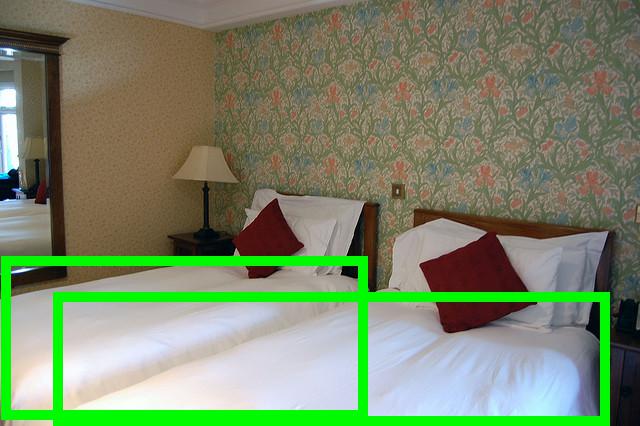}}~~
		\subfigure{\includegraphics[height=40.5mm]{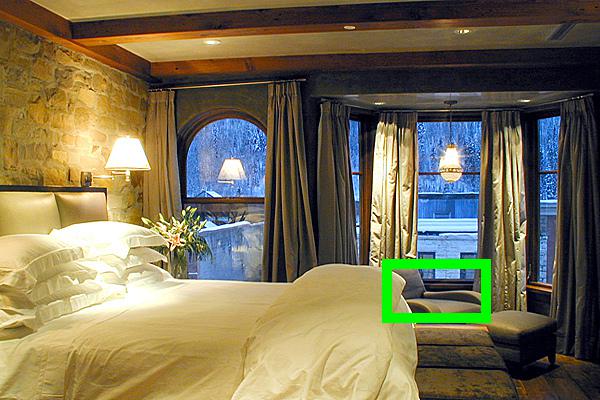}}~~
		\subfigure{\includegraphics[height=40.5mm]{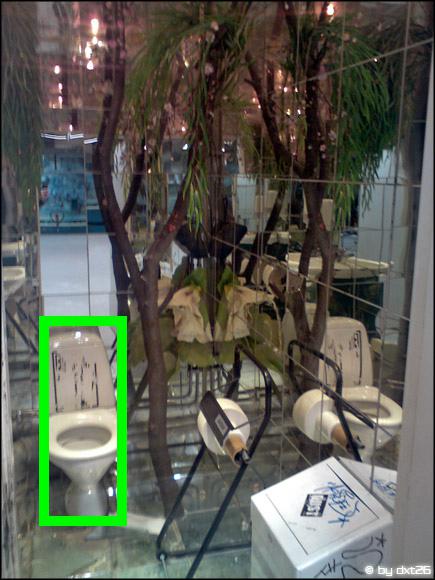}}~~\\
		\caption{\label{fig:anno} Whether an object should be chosen for a task depends on the object properties as well as the presence of better alternatives. The image in the first row shows the objects (green bounding boxes) that have been chosen by the majority of the annotators for the task \texttt{step on something to reach top of a shelf}. While in the first image the table is chosen, the chair is preferred instead of the table in the second image. In the last image, one of the two beds is chosen. The second row shows examples for the task \texttt{sit comfortably}. In the first image, both beds are selected. In the second image, the comfortable chair is preferred over the bed and and the stool. In the third image, the real toilet is selected. Note that the reflection of the toilet in the mirror is annotated as object in COCO as object, but it is not selected since one cannot sit on it.
		}
		\label{fig:dataset_examples}
	\end{figure*}

	In this work, we address the problem of \textit{task driven object detection}. It requires to detect all objects in an image which serve a given task best.
	To this end, we propose the \textit{task driven object detection} (COCO-Tasks) dataset, which is based on the images and annotated objects of the COCO dataset \cite{lin2014microsoft}. For evaluation, we define 14 tasks and asked humans to mark all objects in an image which they favor to solve a given task. If none of the objects in an image is suitable, the annotators were allowed to select none of the objects. The dataset comprises about 40,000 annotated images and for each task between 1,100 and 9,900 objects have been marked by the annotators, where the number of different object categories varies between 6 and 30 for the different tasks.
	Figures~\ref{fig:teaser} and \ref{fig:dataset_examples} show a few examples.
	
	In our experimental evaluation, we show that \textit{task driven object detection} cannot be treated as a standard object detection task. If a standard object detector is trained for each task using the human annotations as ground-truth, the predictions are not very accurate since the favored objects strongly depend on the presence of other objects and their properties. We therefore propose a method based on Gated Graph Neural Networks (GGNN) \cite{ggnn} that explicitly incorporates all detection hypotheses in an image to infer which objects are preferred for a task. Our experimental results show that our proposed method outperforms various ranking and classification based baselines and a thorough ablation study analyzes the design choices of our proposed approach.
	COCO-Tasks dataset and the code for reproducing our experiments are available online\footnote{\href{https://coco-tasks.github.io/}{coco-tasks.github.io}}.
	
	\section{Related Work}
	
	Due to public benchmarks like Pascal VOC \cite{Pascal} and COCO \cite{lin2014microsoft}, there was a tremendous advancement in the area of object detection. State-of-the-art object detectors~\cite{DCN,Soft-NMS,DCN,Chollet2017,Xie2017,Mao2017,GCN,Zhou_2018_CVPR} rely exclusively on convolutional neural networks where in particular Faster R-CNN \cite{faster-RCNN} has been widely used.
	For applications where runtime is critical, other detectors like \cite{YOLO,SSD} provide a very good trade-off between efficiency and accuracy.
	
	In contrast to standard object detection, task driven object detection requires an understanding of the entire scene. This relates it to the task of visual question answering which takes as input a question regarding the content of an image and returns an answer in text form, whereas for task driven object detection the input is a task and the output are bounding boxes around objects that are best suitable for solving the task.
	While~\cite{Antol2015,Gao2015,Yu2015,Ren2015,Malinowski2014} pioneered in visual question answering, \cite{Teney_2018_CVPR, Teney_2018_ECCV, Anderson_2018_CVPR, Nguyen_2018_CVPR, zhang2018learning} are examples of current state-of-the-art methods.
	
	Choosing the best object among the available requires not only recognizing its class but judging its functional attributes, \ie its affordances.
	Detecting and segmenting affordances in images has therefore received an increased interest~\cite{Nguyen2016,Nguyen2017,Kumra2017,cvpr,Do2017}.
	In the work \cite{zhu2015understanding}, learning functional and physical properties together with the handling of objects as tools is investigated. The model is learned from human demonstration and relies on 3d models of objects. The model is then used to recognize tools and affordance regions for 3D objects.
	Fang et al.~\cite{Fang_2018_CVPR} propose to learn to detect affordances from demo videos.
	
	Applying deep neural network on graph structured data has seen a lot of attention from the community recently \cite{graph-1, graph-2, graphconv, ggnn}. Many computer vision problems including scene context can naturally be represented as a graph. Wang and Gupta \cite{video-graph} use a Graph Convolutional Network \cite{graphconv} to represent a video and achieve very good results on video classification. Qi et al.~\cite{graph-seg} have used graph neural networks for semantic segmentation.
	Chuang et al. \cite{Chuang2018Learning} used Gated Graph Neural Networks \cite{ggnn} to model affordances in context. While our work compares the objects to each other, \cite{Chuang2018Learning} focuses on the interaction of objects with their environment.
	
	The task of scene graph generation proposed by Johnson et al.~\cite{Johnson2015Image} requires the detection of objects and relationships between pairs of them. These relationships are typically prepositions indicating relative geometric position and physical interactions.
	While earlier approaches \cite{lu2016visual, zhuang2017towards, peyre2017weakly, zhang2017ppr, zhang2017visual, xu2017scene, li2017vip, liang2017deep, dai2017detecting, li2017scene, newell2017pixels, Zellers2018Neural} avoid the search over the exhaustive number of relations by heuristics, more recently \cite{Yang2018Graph} propose a method which learns to prune unlikely object relationships. While Li et al.~\cite{Li2018Factorizable} rely on modeling subgraphs for scene graph generation,
	Zellers et al.~\cite{Zellers2018Neural} focus on correlations between objects and higher order graph structure statistics.
	
	\begin{figure}
		\includegraphics[trim={7mm, 0, 20mm, 0}, clip, height=46mm]{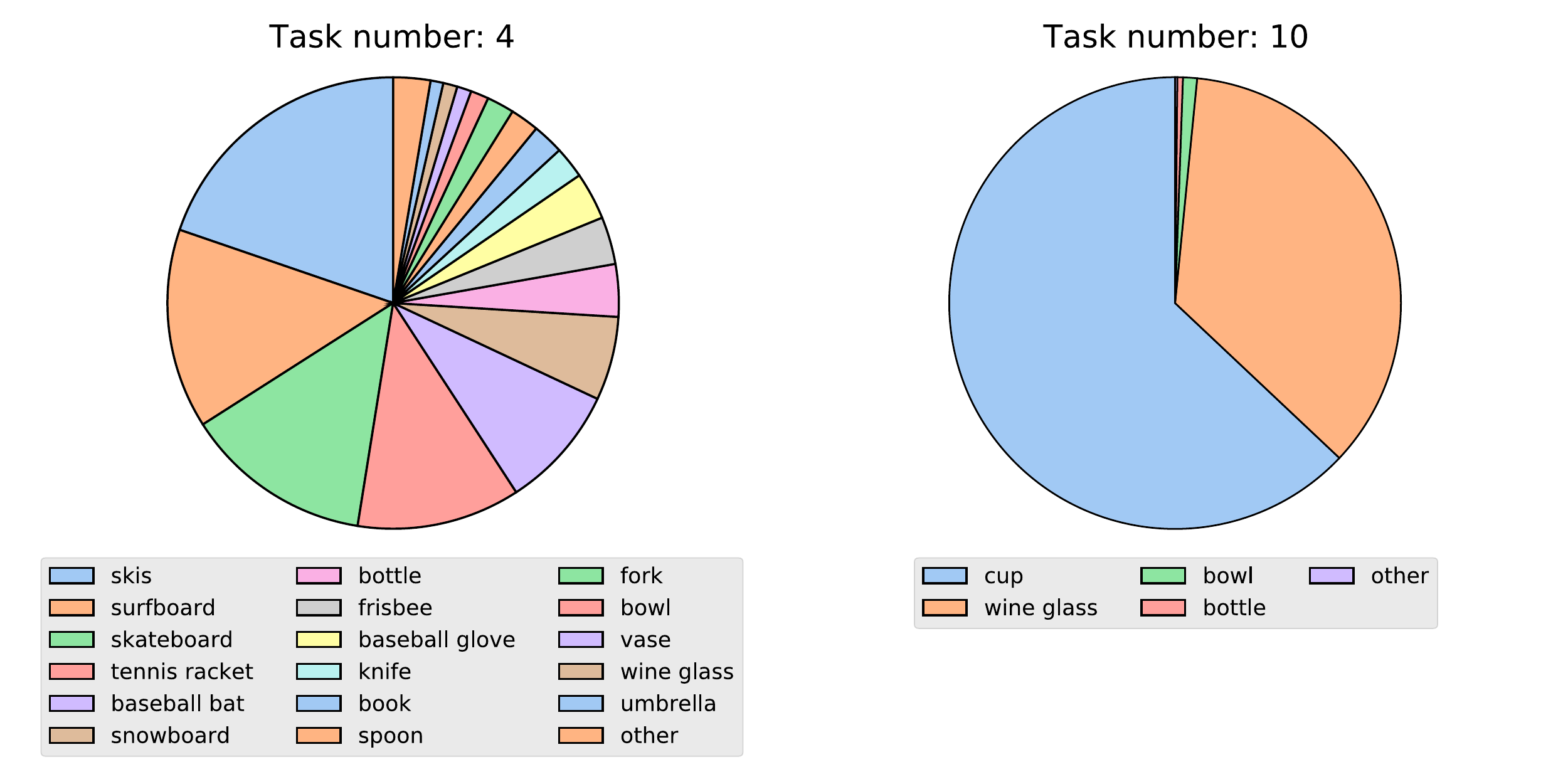}
		\caption{Distribution of chosen objects for task 4 and task 10 across COCO categories. These are the tasks with the highest and lowest number of selected categories, respectively.}
		\label{fig:pie_chart}
	\end{figure}
	
	\begin{table*}
		\scriptsize
		\begin{tabular}{|>{\centering}p{4mm}|p{24mm}|>{\centering}p{20mm}|>{\centering}p{20mm}|>{\centering}p{15mm}|>{\centering}p{15mm}|>{\centering\arraybackslash}p{15mm}|}
			\hline
			& Task & Selected object categories & Objects of all selected categories & Objects chosen by humans & Intra class differentiations & Annotation consistency \\ \hline \hline
			1 &step on something & 12 & 30214 & 5783 &964& 0.927\\ \hline
			2 & sit comfortably& 12 & 31392 & 9870 &1004& 0.938\\ \hline
			3 & place flowers & 10  & 14732 & 3737 &734& 0.925\\ \hline
			4 & get potatoes out of fire & 30  & 32775 & 6889 &525& 0.921\\ \hline
			5 & water plant & 13 &  19050& 4043 & 760& 0.918\\ \hline
			6 & get lemon out of tea& 15 & 22386 & 4707 & 661&0.873\\ \hline
			7 & dig hole & 29& 34015& 6857 &402& 0.922\\ \hline
			8 & open bottle of beer& 12 &  18177 & 1105 & 373& 0.921\\ \hline
			9 & open parcel& 7&  7172& 1759 & 160&0.921\\ \hline
			10 & serve wine& 6&  19209& 3778 & 566&0.963\\ \hline
			11 & pour sugar& 11& 20596& 5739 & 944&0.863\\ \hline
			12 & smear butter & 9&  17489& 1819 & 270&0.896\\ \hline
			13 & extinguish fire& 8&  14821& 2535 & 272&0.940\\ \hline
			14 & pound carpet & 14& 34160& 7176 & 432&0.941\\ \hline
		\end{tabular}
		\caption{\label{tab:affordances} List of the 14 tasks in the COCO-Tasks dataset and some statistics. Selected object categories (column 3) are COCO object categories for which there exists at least one instance chosen by the majority of the annotators for a given task. Column 4 reports how many instances of each of the selected categories are in the images. Column 5 provides the numbers of object instances that are chosen for each task. Column 6 counts the number of instances of categories in an image where at least one instance but not all instances of the same category are selected. Examples of such cases are shown in the last column of Figure~\ref{fig:anno}. The last column reports the probability that two annotators agree if an object is preferred or not. Overall, we have a very high annotation consistency.
		}
		\label{tab:dataset_stats}\label{tab:consistency}
	\end{table*}
	
	\section{COCO-Tasks Dataset}\label{sec:dataset}
	
	Detecting the objects, which are favored for a given task, is very difficult. It requires localizing objects as for a standard object detection task, but the preferred objects in an image vary among image and task. Figure~\ref{fig:dataset_examples} shows a few examples for the first task (\texttt{step on something to reach top of a shelf}) that requires to move an object to a shelf in order to step on it and take something from the top of the shelf, which cannot be reached otherwise.
	The first image shows a table which is selected by the annotator since it serves the task.
	In the second image, however, the table is not selected, since a chair which is much handier is also present.
	This constitutes the additional difficulty of task driven object detection compared to object detection: the validity of a detection also depends on the presence of better options which need to be detected and assessed. One needs to understand the scene in order to judge a particular object.
	The third image shows a task specific preference of instances within an object category: The neglected bed on the left hand side looks heavier than the bed on the right hand side. The height of the bed on the right hand side is also sufficient to reach the top of the shelf.
	In this case, the choice is not anymore at the object category level, but on a finer level where attributes of the instances need to be compared. In summary, task driven object detection requires a detailed understanding of an image, \ie, it needs to be known what objects are in the image and what are the attributes or properties of an instance relative to other objects in an image.
	
	In order to address the problem of task driven object detection, we introduce the COCO-Tasks dataset and we propose a first approach for task driven object detection which will be described
	in Section~\ref{sec:method}.
	The COCO-Tasks dataset is based on the COCO dataset~\cite{lin2014microsoft}, which is the standard benchmark for object detection.
	We have defined 14 tasks which are listed in Table~\ref{tab:affordances} together with some statistics.
	The tasks are quite diverse and include tasks that
	prefer a specific object shape and material like \texttt{serve wine} or \texttt{place flowers} and tasks that are related but require different attributes of
	the objects like \texttt{step on something to reach top of the shelf} or \texttt{sit comfortably}.
	For each of these tasks, we sample 3600 train images from the COCO train2014 split and 900 test images from the COCO val2014 split. To focus on more complex scenes with multiple objects to choose from, we bias the sampling procedure.
	For each task, we define which COCO supercategories are most useful. The list of supercategories per task is provided in the supplementary materials. Then we make sure that 40\% of the images contain multiple categories from these supercategories, 40\% contain exactly one category from these supercategories but multiple instances of it, and 10\% of the images contain exactly one instance from one category. The remaining 10\% are randomly sampled. In total, our train set contains 30,229 images and our test set contains 9,495 images.
	
	\begin{figure}
		\includegraphics[trim={6mm, 0, 0mm, 0}, clip, height=48mm]{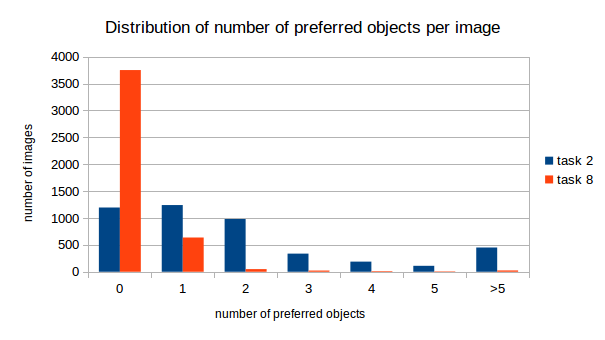}
		\caption{Distribution of the number of preferred objects per image for tasks 2 and 8, which are the tasks with highest and lowest number of selected instances per image, respectively. }
		\label{fig:histogram}
	\end{figure}
	
	In order to annotate the preferred objects in each of the 4,500 images for each task, we use the available COCO segmentation masks.
	We highlight the segmentation masks of all objects annotated in the COCO dataset for the annotator. To specify the requirements for the task on a more intuitive level, we visualize all tasks besides of providing a textual description of the task.
	For instance, we show an image of a shelf for the task \texttt{step on something to reach top of the shelf}.
	The annotators could choose any object, multiple objects or none of them if none of the objects is considered as suitable for this task. The annotators neither knew the procedure of sampling the images nor the supercategories, i.e., they could choose from all 80 COCO categories for each task.
	Each task was annotated by 5 trained annotators. An object is considered to be preferred if it was chosen by the majority of the annotators. Some example annotations are shown in Figure~\ref{fig:dataset_examples}. More information about the annotation tool is provided in the supplementary material.
	
	Table~\ref{tab:dataset_stats} provides some statistics of our dataset. We measured how diverse the selected objects with respect to COCO categories are by counting all categories where at least one instance was selected by the majority of the annotators. The datasets shows a high variation in terms of categories per tasks and the number of selected object categories varies between 6 and 30 depending on the task. From the 80 COCO object class categories of the object detection challenge 2014, instances of 49 classes have been selected for at least one of the 14 tasks. Note that COCO classes also include animals, which are not relevant for the tasks in our dataset.
	We then measured how many instances of all selected categories for each task are present in our datasets, which also largely varies between 7,172 and 34,160 instances. This shows that just reducing the number of categories to a small set that could be relevant for a task would still leave many instances to choose from. The number of instances that have been selected for each task varies between 1,105 and 9,870. We finally provide the number of instances where the annotators differentiate between instances of the same category as it is shown in the last column of Figure~\ref{fig:anno}. In such cases, the properties or attributes of the instances are relevant to make the decision which object should be used.
	In Figure~\ref{fig:pie_chart}, we show the distribution of selected object categories for the tasks with the lowest and highest number of selected object categories. While for \texttt{serving wine} instances from the categories wine glass and cup are mostly selected, there is a large diversity of categories that have been selected for \texttt{getting potatoes out of the fire}.
	Additionally, we report the distribution of the number of selected instances per image in Figure~\ref{fig:histogram}.
	While for \texttt{open bottle of beer} the number of suitable objects is low, there is large diversity in the number of selected instances per image for \texttt{sitting comfortable}. The examples show the large variety of category and instance distributions among the tasks. Additional plots are provided in the supplementary material.
	Furthermore, we evaluated the consistency of the annotations. For each task and each object, we calculated the probability that two annotators agree if this object is preferred or not. As can be seen from Table~\ref{tab:consistency}, the consistency between annotators is very high.
	
	As evaluation metric, we use the AP@0.5 object detection evaluation metric of the COCO detection challenge \cite{lin2014microsoft} where
	the preferred objects for a particular task are the ground truth instances to calculate average precision on. Taking the mean over the tasks yields mAP@0.5.
	
	\section{Task Driven Object Detection}\label{sec:method}
	
	\begin{figure*}
		\subfigure{\includegraphics[height=45mm]{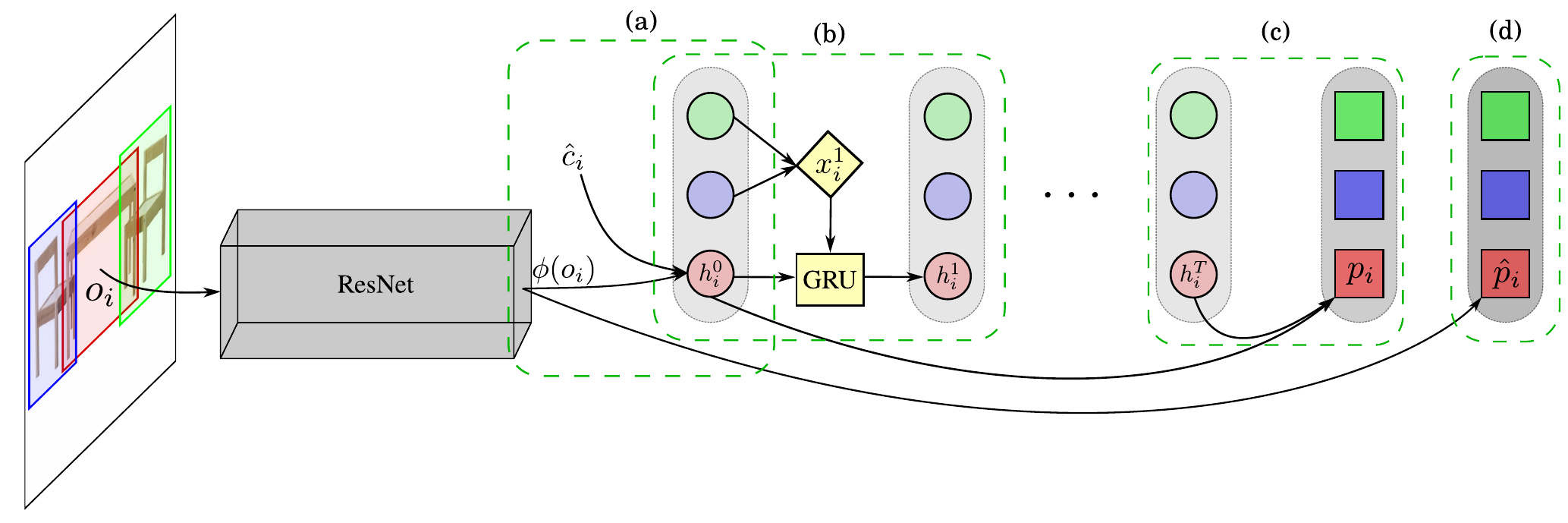}}
		\caption{Overview of the proposed method. Given an image containing a number of objects (3 in this example shown with colors green, blue and red), our method first extracts ResNet features from each bounding box containing the object. (a) Using the extracted features and one-hot encoding of the detected class of the object ($\hat{c}_i$), we compute the initial hidden state of the graph node corresponding to that object using \eqref{ggnn-init}. (b) Using the hidden states of all of the other graph nodes, we aggregate the scene information using \eqref{ggnn-agg} and update the node's hidden state as in \eqref{ggnn-update}. (c) After $T$ iterations of the GGNN, we combine each node's initial and final hidden states using \eqref{ggnn-output} to compute the probability of that object being favored for the task. (d) Finally to make the features learned by the ResNet discriminative we also force the network to estimate suitability scores only from visual features of a single object. At test time, we average the two estimated probabilities.
		}
		\label{fig:method}
	\end{figure*}
	
	In order to identify the most suitable objects in an image for a task, it is required to understand what objects are in the scene and why is an object preferred to other present objects. While the objects in an image can be detected by an off-the-shelf object detector, we have to model the relations of all present objects in an image to select the preferred objects among all detected objects. To this end, we will use a Gated Graph Neural Network (GGNN) \cite{ggnn} to model the global information of all objects in an image.

	\subsection{Proposed Method}\label{sec:proposed_method}
	Our model consists of a ResNet101 \cite{resnet} network without the final fully connected layer with the weights initialized from ILSVRC.
	On top of the ResNet features, we construct a Gated Graph Neural Network \cite{ggnn} where each node is an object in the image and each node is connected to all of the other nodes to gather the information from all of the objects present in the scene.
	On top of the GGNN, we have a fully connected layer which predicts the probability of each object being suitable for each task. We train the whole network end-to-end using binary cross entropy loss. Below we will describe the model in more detail.
	
	An overview of our method is shown in Figure \ref{fig:method}. Given an input image $I$ and a collection of $N$ detected objects in that image $o_{i}, i = 1, ..., N$ specified with their corresponding bounding boxes $b_i$, detection scores $d_i$ and predicted category $c_i$, our method predicts $p_{i}$ the probability of the object $o_i$ being selected for a task.
	
	We first preprocess the bounding boxes by making them square and 10\% larger in each dimension, and then crop the image with the preprocessed bounding boxes. We then extract the features from each cropped bounding box arriving at $\phi(o_i)$.
	
	We create a GGNN with one node for each object in the image. We set the initial hidden value of each node based on the one-hot encoding of the category of that object $\hat{c}_i$ and the ResNet features $\phi(o_i)$ such that
	\begin{equation}
		\label{ggnn-init}
		h_{i}^{0} = g(W_c \hat{c}_i) \odot g(W_\phi \phi(o_i))
	\end{equation}
	where $g(.)$ is the ReLU activation, $\odot$ is the element-wise multiplication and $W_c$ and $W_\phi$ are parameters of the model.
	
	At each step of the GGNN, we first aggregate the information from all other nodes in the graph:
	\begin{equation}
		\label{ggnn-agg}
		x_{i}^{t} = \sum_{j, j \ne i} W_p d_{j} h_{j}^{t-1} + b_p
	\end{equation}
	where $W_p$ and $b_p$ are the parameters of the learned linear mapping in the aggregation step. This corresponds to a graph where each node is connected to all other nodes. We call the multiplication of $d_j$ in \eqref{ggnn-agg} \textit{weighted aggregation}. It gives the possibility to our method to account for misinformation in bad detections with low detection scores.
	Using the aggregated $x_{i}^{t}$ and the previous hidden state of the node $h_{i}^{t-1}$ we arrive at the new hidden state of each node in the graph using the GRU \cite{gru} update rule
	\begin{align}
		z_{i}^{t} =& \sigma(W_z x_{i}^{t} + U_z h_{i}^{t-1} + b_z) \nonumber \\
		r_{i}^{t} =& \sigma(W_r x_{i}^{t} + U_r h_{i}^{t-1} + b_r) \nonumber \\
		\hat{h}_{i}^{t} =& \mathnormal{tanh}(W_h x_{i}^{t} + U_h (r_{i}^{t} \odot h_{i}^{t-1}) + b_h) \nonumber \\
		h_{i}^{t} =& (1 - z_{i}^{t}) \odot h_{i}^{t - 1} + z_{i}^{t} \odot \hat{h}_{i}^{t} \label{ggnn-update}
	\end{align}
	where $\sigma$ is the sigmoid activation and the GRU weights ($W_z$, $W_r$, $W_h$, $U_z$, $U_r$, $U_h$, $b_z$, $b_r$, $b_h$) are learned end-to-end and are shared between all tasks just like the ResNet backbone network. This update rule is applied $T$ times. In our experiments $T$ is set to 3.
	We observed that increasing $T$ does not improve our results.
	
	At the end of the $T$ iterations the model calculates the probability estimate from the concatenation of the initial and final hidden state of each node
	\begin{equation}
		\label{ggnn-output}
		p_{i} = \sigma( f([h_{i}^{0} ; h_{i}^{T}] ))
	\end{equation}
	while learning the weights. $f(.)$ corresponds to a 2 layer fully connected MLP with ReLU activations for the hidden layer where the final layer has a single output. We can modify this output model to generate one probability value for each task using a final layer with $M$ outputs where $M$ is the number of tasks and train a single model for all tasks jointly.
	In order to make the features learned by the ResNet discriminative, we also directly compute suitability estimates
	\begin{equation}
		\label{ggnn-output-aux}
		\hat{p}_{i} = \sigma( \hat{f}(\phi(o_i) ))
	\end{equation}
	from only ResNet features $\phi(o_i)$ as shown in Figure \ref{fig:method} (d). We use two binary cross entropy losses during training for $p_i$ and $\hat{p}_i$. At test time, we use average fusion of $p_i$ and $\hat{p}_i$ to estimate the final probability.
	
	To train our model, we construct each minibatch from objects inside a single image from our training set. All COCO annotated objects are included in the batch, the ones which are specified by our dataset as being preferred for a task are considered as positive examples for that task and the others are considered as negative. Since we use the COCO annotated bounding boxes during training, we set all $d_i$s to 1. During testing, we first perform standard object detection on the test image and get a set of object bounding boxes and their corresponding detection scores and categories. We then perform testing by constructing a batch from all of the detected objects and estimate the probability of each object being preferred for each task as described above.
	The final confidence for mAP evaluation is obtained by multiplying the detection score with the estimated probability.
	Implementation details are provided in the supplementary material.
	
	\section{Experiments}
	In this section, we first evaluate the performance of several baselines as well as our proposed method on COCO-Tasks. After that, we demonstrate both qualitatively and quantitatively that our proposed method learns useful information about the scene context. Furthermore with ablation experiments, we show the benefits of each component of our proposed method. For all of our experiments except the object detection baseline we train and test the models three times and report the average performance numbers.
	
	\subsection{Comparison to Baselines}
	\begin{table}
		\centering
		\scriptsize
		\begin{tabular}{|c|c|c|c|}
			\hline
			\multicolumn{4}{|c|}{Comparison to Baselines mAP@0.5} \\ \hline
			& gt bbox & Faster-RCNN detections & Yolo detections\\ \hline
			object detector & - & 0.206 & - \\ \hline
			pick best class & 0.386 & 0.141 & -\\ \hline
			ranker& 0.564 & 0.091 & -\\ \hline
			classification & 0.616 & 0.288 & 0.291\\ \hline
			proposed + fusion & \textbf{0.742} & \textbf{0.326} & \textbf{0.332}\\ \hline
		\end{tabular}
		\caption{Comparison of the proposed method to several baselines on ground truth bounding boxes as well as Faster-RCNN \cite{faster-RCNN} detections.
			The classification baseline is the strongest one but achieves 12.6\% lower mAP on ground truth bounding boxes and 3.8\% lower mAP on detections compared for our proposed approach.}
		\label{tab:baselines}
	\end{table}
	
	For the object detection baseline, we train a separate object detector for each task on our train set and infer on the test set.
	For all other baselines as well as the proposed method, we train the respective method on ground truth bounding boxes of all COCO objects in the train set.
	We then evaluate all algorithms on (a) ground truth bounding boxes of COCO objects and (b) COCO object detections of a Faster-RCNN object detector \cite{faster-RCNN}.
	While the latter evaluates the performance in a realistic scenario, the former demonstrates the potential of our method that can be reached with a perfect object detector.
	As metric, we use mAP@0.5 for all experiments and report the numbers in Table~\ref{tab:baselines}.
	
	\textbf{Object Detector Baseline.}
	The most straightforward approach for task driven object detection is to treat it as a standard object detection task. To this end, for each of the 14 tasks, we train a 1-class object detector.
	All objects preferred for the respective task constitute the object class to detect. As detector, we use the same Faster-RCNN implementation. Apart from changing the number of classes from 80 to 1, we
	reduce the learning rate from 0.005 to 0.0001, all other hyperparameters stay identical. As reported in Table~\ref{tab:baselines}, this yields an mAP@0.5 of 20.6\%, which is more than 10\% lower than the proposed approach.
	This verifies that task driven object detection can not be treated as a standard object detection task because of the necessity to look for scene context and all present objects.
	
	\textbf{Pick Best Class Baseline.}
	COCO classes differ significantly in their suitability for household tasks. To analyse this effect, we first rank the classes for each task by the fraction of all instances of this class to be preferred on the train set.
	Then for each task and each image of the test set, we omit all detections with detection confidence lower than 0.1. Among the remaining detections, we determine the highest ranked class and only keep the detections belonging to this class with their detection confidence as final confidence.
	The result is 14.1\% on detections which is significantly worse than the \textit{object detector} baseline. On ground truth bounding boxes, this baseline yields only 38.6\% mAP@0.5. This shows that the task driven object detection problem is not solvable by
	the category information alone, but visual information from the objects and image context are required.
	
	\textbf{Ranker Baseline.}
	For the \textit{ranker} baseline, we train a model similar to Deep Relative Attributes \cite{Souri2016} to rank COCO objects in terms of their suitability for a task.
	We exchange the original VGG16 backbone \cite{VGG} of the ranker for a ResNet101 \cite{resnet} backbone to make the method comparable to other baselines. We train a model for each task separately using the Adam optimizer with $10^{-4}$ learning rate for 3 epochs to remain as close as possible to \cite{Souri2016}.
	As for the \textit{pick best class} baseline, we prefilter the detections by a detection confidence threshold of 0.1. Then for each image and each task, we rank all $n$ detections and assign each detection $i$ of rank $r_{i}$ the confidence $c_{i}=1-\frac{r_{i} - 1}{n}$.
	Although on ground truth bounding boxes this method performs better than \textit{pick best class}, it is the worst baseline on detections giving only 9.1\% mAP@0.5 as can be seen from Table~\ref{tab:baselines}. The reason is that a single detection ranked erroneously highly affects all other detections.
	
	\textbf{Classification Baseline.}
	To investigate if a global analysis of all objects present in the scene is necessary, we train a binary classifier on top of the ResNet features for each task and apply it on detections and ground truth bounding boxes.
	As for the proposed method, we obtained the final confidence by multiplying the classifier output and the detector confidence.
	This baseline model is equivalent to our method, without the class information input, the context modeling using a graph and joint training for all tasks simultaneously.
	This is the strongest baseline as can be seen from Table~\ref{tab:baselines}. It gives 61.6\% on ground truth bounding boxes and 28.8\% on detections. However, this is still substantially below our proposed method which takes the scene context into account.
	
	\textbf{Proposed Method.}
	The proposed method with fusion where we average $p_i$ and $\hat{p}_i$ for final estimate, yields 32.6\% on Faster-RCNN \cite{faster-RCNN} detections and 74.2\% on ground truth bounding boxes outperforming our baselines by a large margin. Various ablation experiments showing the effect of different components of our method will follow.
	
	\textbf{Other Detector}
	Our method outperforms the strongest classifier baseline even if we use the Yolov2 detector \cite{YOLO} as can be seen from Table~\ref{tab:baselines}.
	\subsection{Ablation Experiments} \label{sec:ablation}
	
	\begin{table}
		\centering
		\scriptsize
		\begin{tabular}{|c|c|c|}
			\hline
			\multicolumn{3}{|c|}{Ablation experiment results, mAP@0.5} \\ \hline
			& gt bbox & Faster-RCNN detections\\ \hline
			classifier & 0.616 & 0.288\\ \hline
			(a) joint classifier & 0.647 & 0.302 \\ \hline
			(b) joint classifier + class & 0.719 & 0.301\\ \hline
			(c) joint GGNN + class & 0.763 & 0.293\\ \hline
			(d) joint GGNN + class + w. aggreg. & - & 0.303\\ \hline
			(e) proposed & \textbf{0.771}& 0.318\\ \hline
			(f) proposed + fusion & 0.742 & \textbf{0.326}\\ \hline
			(g) no visual input & 0.589 & 0.237\\ \hline
			(h) no visual input + bounding box & 0.412 & 0.152\\ \hline
		\end{tabular}
		\caption{Evaluation of the components of our proposed method. We start with a task wise classifier, (a) then add joint training, (b) add COCO classes as input, (c) introduce the GGNN, (d) add weighted aggregation, (e) add the discriminatory loss and (f) perform fusion. Further ablation experiments (g) and (h) reveal the impact of the visual information.}
		\label{tab:ablations}
	\end{table}

	We observed that the classification baseline was lacking in performance compared to our proposed method. This is due to the differences between the classification baseline and our proposed method. These differences are: (a) joint training of all tasks together, (b) direct class information input, and (c) GGNN for scene context modeling. We will add these 3 components one by one to the classification baseline and show the effect of each of them. Furthermore, in our GGNN we show the effect of (d) \textit{weighted aggregation}, (e) the direct discriminatory loss on top of the ResNet features and (f) fusion of $p_i$ and $\hat{p}_i$ for the final probability estimate.
	The results for these ablation experiments are reported in Table~\ref{tab:ablations}.
	
	\textbf{a) Joint Training.}
	While for the classification baseline we train a separate classifier for each of the tasks, a first improvement can be easily obtained by training a classifier jointly for all tasks, \ie using shared features. This is done by replacing the final single output fully connected layer that estimates $p_i$ into a layer with $M$ outputs, where $M$ is the number of tasks.
	If a task is annotated for an image during training, we calculate the binary cross entropy loss and skip that task otherwise.
	Training the classifier jointly increases the performance on ground truth bounding boxes from 61.6\% to 64.7\% and on detections from 28.8\% to 30.2\%. We think this is due to the higher number of training images and better features that are learned by ResNet.
	
	\textbf{b) Direct Class Information Input.}
	The object's class as a direct input provides additional valuable information that might be harder for the network to learn from ResNet features.
	Given this insight we use \eqref{ggnn-init} to combine the ResNet features ($\phi(o_i)$) and the one-hot encoding of the classes ($\hat{c}_i$) as it is done in our proposed method. We then use the hidden representations $h_i^0$ as input to the final classification layer. During training we use the ground truth class, during inference we use the detected classes which might be noisy.
	On ground truth, the results get boosted from 64.7\% to 71.9\%. However, on detections, the performance stays almost the same. We reckon that this is due to the difference between reliable ground truth classes during training and erroneous classes as predicted by the detector during inference. In our proposed method, this problem is addressed by our \textit{weighted aggregation} mechanism.
	
	\textbf{c) GGNN for Scene Context Modeling.}
	We now add the GGNNs as described in Section~\ref{sec:proposed_method} to see the effect of scene context modeling. For this ablation experiment,
	the \textit{weighted aggregation} (by setting all $d_i$s in \eqref{ggnn-agg} to 1) and the discriminator loss are not used.
	This is equivalent to a simplified GGNN.
	On ground truth bounding boxes we get an improvement of 4.4\% arriving at 76.3\% as a result of scene context modeling, but on detections the performance slightly drops to 29.3\%. The detection confidence problems encountered by the classifier are amplified, since the GGNN takes all detections into account when judging a single one. Thus the final result for each detection is affected by low confidence detections during inference. Typically these are wrong detections, thus the GGNN is confronted with visual input not seen during training. To solve this issue, we have incorporated the \textit{weighted aggregation}.
	
	\textbf{d) Weighted Aggregation.}
	By the \textit{weighted aggregation}, we take the confidences of the detections $d_i$s into account \eqref{ggnn-agg}.
	We observe that addition of such weighting improves our results considerably on detections.
	This thwarts the propagation of visual features of low confidence detections through the GGNN resulting in an improvement from 29.3\% to 30.3\%.
	Note that the \textit{weighted aggregation} does not change the result on ground truth bounding boxes since the $d_i$s are equal to 1 in this case.
	
	\textbf{e) Direct Discriminator Loss.}
	We also impose intermediate supervision on the visual features fed into the initializer. We add a fully connected layer mapping these features onto probabilities for each task and apply a task wise binary cross entropy loss to these probabilities. This loss makes the visual features more discriminative for the final goal. The features give a better backup in case the class information is not correct. In general such a loss improves the performance of our model to 77.1\% and 31.8\% on ground truth bounding boxes and detections, respectively.
	
	\textbf{f) Probability Fusion.}
	Average fusion of the probabilities $p_i$ from (\ref{ggnn-output}) and $\hat{p}_i$ from (\ref{ggnn-output-aux}) further improves the results on detections. We observe that this does cause some performance decrease for the case of a perfect detector.
	The fusion is therefore only relevant if the detections are noisy.
	
	\textbf{g) Removing Visual Input $\boldsymbol{\phi(o_i)}$.}
	Since class information improves the results on ground truth bounding boxes significantly, the question comes to mind if visual information inside the bounding boxes is necessary at all.
	To test this, we do not use the visual features $\phi(o_i)$ for GGNN and only keep the class information as input.
	As a result, the mAP significantly drops, showing that the appearance of the objects is very important for the task and that GGNN takes it into account.
	
	\textbf{h) Removing Visual Input $\boldsymbol{\phi(o_i)}$ and Adding Bounding Box Geometry.}
	We then used the coordinates of the bounding boxes normalized by image width and height $b(o_i)$ instead of the visual features $\phi(o_i)$ for GGNN (proposed no vis. input + bbox).
	This leads to even worse results since the model overfits to the coordinates of the bounding boxes of the objects inside the training images.
	
	In the supplementary material we provide the results for each task.
	
	\subsection{Scene Context Learned by GGNN}
	
	\begin{figure}
		\subfigure{\includegraphics[trim={10mm, 0, 0, 0}, clip, height=42mm]{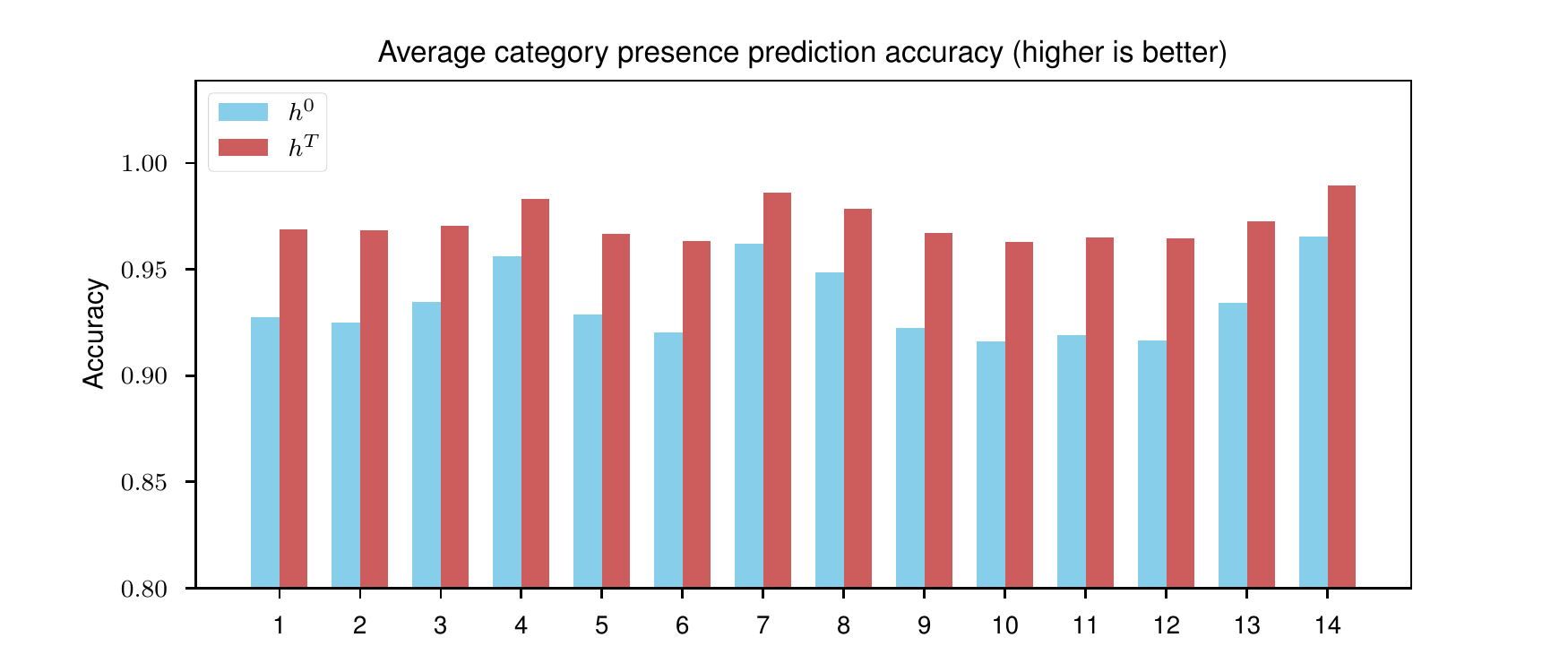}}
		\caption{
			Accuracy of the predicted categories when retrieving nearest neighbors based on $h_i^{0}$ vs. retrieving based on $h_i^{T}$.
		}
		\label{fig:scene_count}
	\end{figure}
	
	The aim of introducing the GGNN was to consider scene context in our model.
	Intuitively, the GGNN aggregates the information about all objects relevant for the task which are present in the image and stores them in the final hidden node representation $h_i^T$.
	To prove this intuition quantitatively, we retrieve the 5 most similar objects for each task and each object of the test set. Then we use the categories present in the scene of the query object as a prediction for the categories present in the scene of the retrieved objects and measure the prediction accuracy. We compute the similarity based on $h_i^T$, which should contain scene information and compare it to the similarity computed based on $h_i^0$.
	
	The prediction accuracies are high in both cases, which is primarily due to the fact that most COCO categories are absent in any image. However as can be seen from Figure~\ref{fig:scene_count}, when retrieving based on similarity of $h_i^{0}$ instead of $h_i^{T}$, the accuracy of this prediction is significantly lower for all tasks. This verifies our intuition. More analysis and qualitative examples are provided in the supplementary material.
	
	\section{Conclusion}
	In this work, we have addressed the problem of task driven object detection. In contrast to standard object detection, it requires to detect and select the best objects for solving a given task. To study this problem, we created a dataset based on the COCO dataset \cite{lin2014microsoft}. It comprises about 40k images with annotations for 14 tasks. We evaluated several baselines based on ranking or classification approaches on this dataset. We furthermore introduced a novel approach for this task that takes as input all detected objects in an image and uses a Gated Graph Neural Network to model the relations of the object hypotheses in order to infer the objects that are preferred for a given task.
	
	\textbf{Acknowledgment}
	This work has been funded by the Deutsche Forschungsgemeinschaft (DFG,
	German Research Foundation) – GA 1927/5-1 (FOR 2535 Anticipating Human
	Behavior) and the ERC Starting Grant ARCA (677650).

	{\small
		\bibliographystyle{ieee}
		\bibliography{ms}

\begin{thebibliography}{10}\itemsep=-1pt

\bibitem{Anderson_2018_CVPR}
Peter Anderson, Xiaodong He, Chris Buehler, Damien Teney, Mark Johnson, Stephen
  Gould, and Lei Zhang.
\newblock Bottom-up and top-down attention for image captioning and visual
  question answering.
\newblock {\em CVPR}, 2018.

\bibitem{Antol2015}
Stanislaw Antol, Aishwarya Agrawal, Jiasen Lu, Margaret Mitchell, Dhruv Batra,
  C Lawrence~Zitnick, and Devi Parikh.
\newblock {VQA}: Visual question answering.
\newblock {\em ICCV}, 2015.

\bibitem{Soft-NMS}
Navaneeth Bodla, Bharat Singh, Rama Chellappa, and Larry~S Davis.
\newblock Soft-{NMS} -- {I}mproving object detection with one line of code.
\newblock {\em ICCV}, 2017.

\bibitem{gru}
Kyunghyun Cho, Bart Van~Merri{\"e}nboer, Dzmitry Bahdanau, and Yoshua Bengio.
\newblock On the properties of neural machine translation: Encoder-decoder
  approaches.
\newblock {\em SSST}, 2014.

\bibitem{Chollet2017}
Fran{\c{c}}ois Chollet.
\newblock Xception: Deep learning with depthwise separable convolutions.
\newblock {\em CVPR}, 2017.

\bibitem{Chuang2018Learning}
Ching-Yao Chuang, Jiaman Li, Antonio Torralba, and Sanja Fidler.
\newblock Learning to act properly: Predicting and explaining affordances from
  images.
\newblock {\em CVPR}, 2018.

\bibitem{dai2017detecting}
Bo Dai, Yuqi Zhang, and Dahua Lin.
\newblock Detecting visual relationships with deep relational networks.
\newblock {\em CVPR}, 2017.

\bibitem{DCN}
Jifeng Dai, Haozhi Qi, Yuwen Xiong, Yi Li, Guodong Zhang, Han Hu, and Yichen
  Wei.
\newblock Deformable convolutional networks.
\newblock {\em ICCV}, 2017.

\bibitem{Do2017}
Thanh-Toan Do, Anh Nguyen, Ian Reid, Darwin~G. Caldwell, and Nikos~G
  Tsagarakis.
\newblock Affordancenet: An end-to-end deep learning approach for object
  affordance detection.
\newblock {\em ICRA}, 2018.

\bibitem{graph-2}
David~K Duvenaud, Dougal Maclaurin, Jorge Iparraguirre, Rafael Bombarell,
  Timothy Hirzel, Alan Aspuru-Guzik, and Ryan~P Adams.
\newblock Convolutional networks on graphs for learning molecular fingerprints.
\newblock In {\em NIPS}. 2015.

\bibitem{Pascal}
Mark Everingham, SM~Ali Eslami, Luc Van~Gool, Christopher~K.I. Williams, John
  Winn, and Andrew Zisserman.
\newblock The {PASCAL} {V}isual {O}bject {C}lasses {C}hallenge 2012: {A}
  {R}etrospective.
\newblock {\em IJCV}, 2014.

\bibitem{Fang_2018_CVPR}
Kuan Fang, Te-Lin Wu, Daniel Yang, Silvio Savarese, and Joseph~J. Lim.
\newblock Demo2{V}ec: Reasoning object affordances from online videos.
\newblock In {\em CVPR}, 2018.

\bibitem{Gao2015}
Haoyuan Gao, Junhua Mao, Jie Zhou, Zhiheng Huang, Lei Wang, and Wei Xu.
\newblock Are you talking to a machine? {D}ataset and methods for multilingual
  image question.
\newblock {\em NIPS}, 2015.

\bibitem{resnet}
Kaiming He, Xiangyu Zhang, Shaoqing Ren, and Jian Sun.
\newblock Deep residual learning for image recognition.
\newblock {\em CVPR}, 2016.

\bibitem{graph-1}
Mikael Henaff, Joan Bruna, and Yann LeCun.
\newblock Deep convolutional networks on graph-structured data.
\newblock {\em arXiv preprint arXiv:1506.05163}, 2015.

\bibitem{Johnson2015Image}
Justin Johnson, Ranjay Krishna, Michael Stark, Li-Jia Li, David Shamma, Michael
  Bernstein, and Li Fei-Fei.
\newblock Image retrieval using scene graphs.
\newblock {\em CVPR}, 2015.

\bibitem{graphconv}
Thomas~N Kipf and Max Welling.
\newblock Semi-supervised classification with graph convolutional networks.
\newblock {\em ICLR}, 2017.

\bibitem{Kumra2017}
Sulabh Kumra and Christopher Kanan.
\newblock Robotic grasp detection using deep convolutional neural networks.
\newblock {\em IROS}, 2017.

\bibitem{li2017vip}
Yikang Li, Wanli Ouyang, and Xiaogang Wang.
\newblock Vi{P}-{CNN}: A visual phrase reasoning convolutional neural network
  for visual relationship detection.
\newblock {\em CVPR}, 2017.

\bibitem{Li2018Factorizable}
Yikang Li, Wanli Ouyang, Bolei Zhou, Jianping Shi, Chao Zhang, and Xiaogang
  Wang.
\newblock Factorizable net: An efficient subgraph-based framework for scene
  graph generation.
\newblock {\em ECCV}, 2018.

\bibitem{li2017scene}
Yikang Li, Wanli Ouyang, Bolei Zhou, Kun Wang, and Xiaogang Wang.
\newblock Scene graph generation from objects, phrases and region captions.
\newblock {\em ICCV}, 2017.

\bibitem{ggnn}
Yujia Li, Daniel Tarlow, Marc Brockschmidt, and Richard Zemel.
\newblock Gated graph sequence neural networks.
\newblock {\em ICLR}, 2016.

\bibitem{liang2017deep}
Xiaodan Liang, Lisa Lee, and Eric~P Xing.
\newblock Deep variation-structured reinforcement learning for visual
  relationship and attribute detection.
\newblock {\em CVPR}, 2017.

\bibitem{lin2014microsoft}
Tsung-Yi Lin, Michael Maire, Serge Belongie, James Hays, Pietro Perona, Deva
  Ramanan, Piotr Doll{\'a}r, and C~Lawrence Zitnick.
\newblock Microsoft coco: Common objects in context.
\newblock In {\em ECCV}. 2014.

\bibitem{SSD}
Wei Liu, Dragomir Anguelov, Dumitru Erhan, Christian Szegedy, Scott Reed,
  Cheng-Yang Fu, and Alexander~C Berg.
\newblock {SSD}: Single shot multibox detector.
\newblock {\em ECCV}, 2016.

\bibitem{lu2016visual}
Cewu Lu, Ranjay Krishna, Michael Bernstein, and Li Fei-Fei.
\newblock Visual relationship detection with language priors.
\newblock {\em ECCV}, 2016.

\bibitem{Malinowski2014}
Mateusz Malinowski and Mario Fritz.
\newblock A multiworld approach to question answering about realworld scenes
  based on uncertain input.
\newblock {\em NIPS}, 2014.

\bibitem{Mao2017}
Jiayuan Mao, Tete Xiao, Yuning Jiang, and Zhimin Cao.
\newblock What can help pedestrian detection?
\newblock {\em CVPR}, 2017.

\bibitem{newell2017pixels}
Alejandro Newell and Jia Deng.
\newblock Pixels to graphs by associative embedding.
\newblock {\em NIPS}, 2017.

\bibitem{Nguyen2016}
Anh Nguyen, Dimitrios Kanoulas, Darwin~G. Caldwell, and Nikos~G. Tsagarakis.
\newblock Detecting object affordances with convolutional neural networks.
\newblock {\em IROS}, 2016.

\bibitem{Nguyen2017}
Anh Nguyen, Dimitrios Kanoulas, Darwin~G Caldwell, and Nikos~G Tsagarakis.
\newblock Object-based affordances detection with convolutional neural networks
  and dense conditional random fields.
\newblock {\em IROS}, 2017.

\bibitem{Nguyen_2018_CVPR}
Duy-Kien Nguyen and Takayuki Okatani.
\newblock Improved fusion of visual and language representations by dense
  symmetric co-attention for visual question answering.
\newblock In {\em CVPR}, 2018.

\bibitem{GCN}
Chao Peng, Xuangyu Zhang, Gang Yu, Guiming Luo, and Jian Sun.
\newblock Large kernel matters--{I}mprove semantic segmentation by global
  convolutional network.
\newblock {\em CVPR}, 2017.

\bibitem{peyre2017weakly}
Julia Peyre, Ivan Laptev, Cordelia Schmid, and Josef Sivic.
\newblock Weakly-supervised learning of visual relations.
\newblock {\em ICCV}, 2017.

\bibitem{graph-seg}
Xiaojuan Qi, Renjie Liao, Jiaya Jia, Sanja Fidler, and Raquel Urtasun.
\newblock 3{D} graph neural networks for {RGBD} semantic segmentation.
\newblock In {\em CVPR}, 2017.

\bibitem{YOLO}
Joseph Redmon and Ali Farhadi.
\newblock {YOLO} 9000: Better, faster, stronger.
\newblock {\em CVPR}, 2017.

\bibitem{Ren2015}
Mengye Ren, Ryan Kiros, and Richard Zemel.
\newblock Exploring models and data for image question answering.
\newblock {\em NIPS}, 2015.

\bibitem{faster-RCNN}
Shaoqing Ren, Kaiming He, Ross Girshik, and Jian Sun.
\newblock Faster {R-CNN}: Towards real-time object detection with region
  proposal networks.
\newblock {\em ICCV}, 2015.

\bibitem{cvpr}
Johann Sawatzky, Abhilash Srikantha, and Juergen Gall.
\newblock Weakly supervised affordance detection.
\newblock {\em CVPR}, 2017.

\bibitem{VGG}
Karen Simonyan and Andrew Zisserman.
\newblock Very deep convolutional networks for large-scale image recognition.
\newblock In {\em ICLR}, 2015.

\bibitem{Souri2016}
Yaser Souri, Erfan Nouri, and Ehsan Adeli.
\newblock Deep relative attributes.
\newblock {\em ACCV}, 2016.

\bibitem{Teney_2018_CVPR}
Damien Teney, Peter Anderson, Xiaodong He, and Anton van~den Hengel.
\newblock Tips and tricks for visual question answering: Learnings from the
  2017 challenge.
\newblock {\em ECCV}, 2018.

\bibitem{Teney_2018_ECCV}
Damien Teney and Anton van~den Hengel.
\newblock Visual question answering as a meta learning task.
\newblock {\em ECCV}, 2018.

\bibitem{video-graph}
Xiaolong Wang and Abhinav Gupta.
\newblock Videos as space-time region graphs.
\newblock In {\em ECCV}, 2018.

\bibitem{Xie2017}
Saining Xie, Ross Girshick, Piotr Dollár, Zhuowen Tu, and Kaiming He.
\newblock Aggregated residual transformations for deep neural networks.
\newblock {\em CVPR}, 2017.

\bibitem{xu2017scene}
Danfei Xu, Yuke Zhu, Christopher~B. Choy, and Li Fei-Fei.
\newblock Scene graph generation by iterative message passing.
\newblock {\em CVPR}, 2017.

\bibitem{Yang2018Graph}
Jianwei Yang, Jiasen Lu, Stefan Lee, Dhruv Batra, and Devi Parikh.
\newblock Graph {R-CNN} for scene graph generation.
\newblock {\em ECCV}, 2018.

\bibitem{Yu2015}
Licheng Yu, Eunbyung Park, Alexander~C. Berg, and Tamara~L. Berg.
\newblock Visual madlibs: Fill in the blank description generation and question
  answering.
\newblock {\em ICCV}, 2015.

\bibitem{Zellers2018Neural}
Rowan Zellers, Mark Yatskar, Sam Thomson, and Yejin Choi.
\newblock Neural motifs: Scene graph parsing with global context.
\newblock {\em CVPR}, 2018.

\bibitem{zhang2017visual}
Hanwang Zhang, Zawlin Kyaw, Shih-Fu Chang, and Tat-Seng Chua.
\newblock Visual translation embedding network for visual relation detection.
\newblock {\em CVPR}, 2017.

\bibitem{zhang2017ppr}
Hanwang Zhang, Zawlin Kyaw, Jinyang Yu, and Shih-Fu Chang.
\newblock {PPR-FCN}: Weakly supervised visual relation detection via parallel
  pairwise {R-FCN}.
\newblock {\em ICCV}, 2017.

\bibitem{zhang2018learning}
Yan Zhang, Jonathon Hare, and Adam Prügel-Bennett.
\newblock Learning to count objects in natural images for visual question
  answering.
\newblock {\em ICLR}, 2018.

\bibitem{Zhou_2018_CVPR}
Peng Zhou, Bingbing Ni, Cong Geng, Jianguo Hu, and Yi Xu.
\newblock Scale-transferrable object detection.
\newblock In {\em CVPR}, 2018.

\bibitem{zhu2015understanding}
Yixin Zhu, Yibiao Zhao, and Song Chun~Zhu.
\newblock Understanding tools: Task-oriented object modeling, learning and
  recognition.
\newblock In {\em CVPR}, 2015.

\bibitem{zhuang2017towards}
Bohan Zhuang, Lingqiao Liu, Chunhua Shen, and Ian Reid.
\newblock Towards context-aware interaction recognition for visual relationship
  detection.
\newblock {\em ICCV}, 2017.

\end{thebibliography}
	}
	
\end{document}